\documentclass[final]{neus2025}
\usepackage{subcaption}
\usepackage{enumitem}

\title[Taxonomic Networks]{Taxonomic Networks: A Representation for Neuro-Symbolic Pairing}

\usepackage{times}
\usepackage{graphicx}

\author{%
 \Name{Zekun Wang} \Email{zekun@gatech.edu}\\
 \addr School of Interactive Computing, Georgia Institute of Technology, Atlanta, GA.
 \AND
 \Name{Ethan L. Haarer} \Email{ehaarer3@gatech.edu}\\
 \addr School of Interactive Computing, Georgia Institute of Technology, Atlanta, GA.%
 \AND
 \Name{Nicki Barari} \Email{nicki.barari@drexel.edu}\\
 \addr College of Computing and Informatics, Drexel University, Philadelphia, PA%
 \AND
 \Name{Christopher J. MacLellan} \Email{cmaclell@gatech.edu}\\
 \addr School of Interactive Computing, Georgia Institute of Technology, Atlanta, GA.
}

\begin{document}

\maketitle

\begin{abstract}
We introduce the concept of a \textbf{neuro-symbolic pair}---neural and symbolic approaches that are linked through a common knowledge representation.
Next, we present \textbf{taxonomic networks}, a type of discrimination network in which nodes represent hierarchically organized taxonomic concepts.
Using this representation, we construct a novel neuro-symbolic pair and evaluate its performance.
We show that our symbolic method learns taxonomic nets more efficiently with less data and compute, while the neural method finds higher-accuracy taxonomic nets when provided with greater resources.
As a neuro-symbolic pair, these approaches can be used interchangeably based on situational needs, with seamless translation between them when necessary.
This work lays the foundation for future systems that more fundamentally integrate neural and symbolic computation.
\end{abstract}


\section{Introduction}

Research on neuro-symbolic AI explores the integration of neural and symbolic methods to combine their complementary strengths and mitigate their respective weaknesses. 
Symbolic AI is characterized by its use of high-level \textit{symbolic} representations that closely correspond to the cognitive symbols humans use \citep{newell1980physical}.
This paradigm emphasizes techniques for explicitly leveraging and manipulating these symbols to support capabilities such as inference and planning.
A widely recognized limitation of symbolic AI is its reliance on knowledge engineering to construct these representations.
While hand-authoring limits scalability, it produces symbols that are explicitly linked to human meanings.
As a result, symbolic systems are often interpretable by design, and when their symbolic knowledge is accurate, their outputs are reliably correct.

Neural AI approaches, in contrast, are predominantly data-driven, relying minimally on knowledge engineering.\footnote{However, much of the progress in neural AI research stems from the development of new neural architectures and data-processing techniques, which could be considered forms of engineered knowledge.}
While this data-driven focus has enabled neural methods to achieve impressive performance and widespread adoption, the correspondence between their learned internal representations (i.e., neurons and their activations) and human meaning is often unclear.
Complicating this further, neural networks typically learn distributed representations \citep{hinton1986learning}, in which higher-level human symbols are encoded across multiple, or even all, internal neurons. 
This characteristic is why neural representations are often referred to as \textit{sub-symbolic}---a single cognitive symbol is typically represented through the collective activation of many neurons.
As a result, neural AI systems are inherently less interpretable than symbolic AI systems.
The absence of internal symbols that correspond with human cognitive symbols, coupled with a lack of explicit mechanisms for symbol manipulation, often leads to unreliable outputs.

Several efforts have sought to bridge these paradigms.
For example, \citet{kautz2022third} reviews six approaches for building neuro-symbolic systems, though his analysis primarily focuses on \textit{combination}, where one approach (neural or symbolic) serves as a sub- or co-routine of the other. 
More recent reviews continue to emphasize the integration of distinct neural and symbolic modules \citep{sarker2022neuro,bhuyan2024neuro}.
While such combined systems are straightforward to construct, they retain the fundamental weaknesses of each component---for instance, the neural component may still suffer from interpretability and reliability issues, while the symbolic component may still depend on hand-authoring.
While combination has its merits, we argue that \textit{unification} approaches that blur the boundaries between the neural and symbolic paradigms deserve further exploration.

To this end, we introduce the concept of \textbf{neuro-symbolic pairs}---neural and symbolic approaches that have linked representations, allowing models to be translated between them.
What makes such a pairing possible is the use of a symbolic representation that can also be instantiated within a neural framework.
Developers can use these pairs to seamlessly switch between different paradigms, selecting the one that best suits their current needs. 
For instance, a developer could use a neural approach to learn a model from a large amount of data, then translate the learned model into a symbolic framework for deployment.

In the following sections, we formalize the concept of neuro-symbolic pairs and outline the requirements for their implementation.
We then propose \textbf{taxonomic networks}, a type of discrimination network where the nodes represent categories that are organized taxonomically, as a novel representation that can serve as the foundation for such a pairing.
Next, we present a neuro-symbolic pair for taxonomic networks and evaluate the distinct performance characteristics of the paired elements.
We conclude with a discussion of broader implications and potential next steps.





\vspace{-10px}
\section{Background}

Our methodology is inspired by recent work on mechanistic interpretability.
\citet{elhage2022toy} explores the concept of \textit{monosemantic} neurons---those that activate exclusively in response to a single feature.
An example is a neuron that activates only when presented with multimodal stimuli representing Halle Berry \citep{kim2018deep}.
As this example suggests, monosemantic neurons often closely correspond to cognitive symbols.
Consequently, neural networks that incorporate these neurons exhibit more symbolic-like behavior and are arguably more interpretable \citep{cunningham2023sparse}.
\citet{elhage2022toy} conduct several experiments to investigate the conditions necessary for learning neural networks with monosemantic neurons.
They explore the phenomenon of \textit{superposition}, where neural networks---particularly smaller ones---compress a larger set of features into a smaller set of neurons.
They hypothesize that neural networks in superposition tend to develop \textit{polysemantic} neurons---which activate in response to multiple features.
Their findings suggest that monosemantic neurons are more likely to occur in larger networks (those with more neurons than features) and in networks that employ techniques such as regularization and sparse coding \citep{cunningham2023sparse} to encourage features to align with individual neural activations.
Under the right conditions, it may be possible to learn neural networks that function like symbolic systems---employing internal representations that more closely correspond to cognitive symbols and offering greater mechanistic interpretability.  

We also draw inspiration from prior work on generative-discriminative classifiers.
\citet{ng2001discriminative} introduce this concept and show that na\"{i}ve Bayes and logistic regression form what they call a \textit{generative-discriminative pair}.
Under certain assumptions, they demonstrate that na\"{i}ve Bayes searches the same hypothesis space as logistic regression---both search for a linear hyperplane in the feature space. 
Furthermore, they derive a formula for translating a given na\"{i}ve Bayes model (the generative model) into a logistic regression model (the discriminative model) that makes identical predictions.\footnote{Each na\"{i}ve Bayes model corresponds to a unique logistic regression model, but the reverse mapping is one-to-many.}
Although the two approaches share the same hypothesis space, they exhibit different performance characteristics.
Na\"{i}ve Bayes learns probabilistic prototypes for each class, and these prototypes only implicitly (via Bayes rule) determine the linear decision boundaries between classes. 
In contrast, logistic regression learns a decision boundary directly, without assuming a specific distributional form for the class prototypes---whereas na\"{i}ve Bayes assumes they follow a normal distribution with independent features.
\citet{ng2001discriminative} further show that while these approaches form a pair, they do not necessarily learn the same models.
They find that na\"{i}ve Bayes converges to its asymptotic performance with substantially less data than logistic regression but that logistic regression ultimately achieves better performance when na\"{i}ve Bayes' assumptions are violated and sufficient data is available.
They conclude by arguing that this generative-discriminative pair allows developers to leverage the strengths of both approaches—using na\"{i}ve Bayes in low-data scenarios and transitioning to logistic regression as more data becomes available.

\vspace{-10px}
\section{Neuro-Symbolic Pairs}

Building on these earlier ideas, we propose the concept of a neuro-symbolic pair.
We define the formation of such a pair as consisting of:
 
\begin{enumerate}[topsep=2pt, partopsep=0pt, itemsep=0pt]
    \item Identifying a representation that can be instantiated within both neural and symbolic terms;
    \item Developing neural and symbolic approaches that operate over this shared representation; and
    \item Defining translation operations that convert a model from one framework (neural or symbolic) into the other.\footnote{While bidirectional translation is desirable, as with generative-discriminative pairs, it may not always be feasible.}
\end{enumerate}

\noindent Based on prior research on mechanistic interpretability, we hypothesize that as neural networks become sparser---with more of their neurons becoming monosemantic---they will increasingly resemble their symbolic counterparts.
In other words, in the limit of increasing sparsity, neural networks may effectively function as symbolic systems.
Although sparse coding-based learning is much more intensive than conventional learning, it often produces better models, even with less data \citep{coates2012learning,hannan2023mobileptx}.
However, even if a neural network becomes functionally symbolic, it would still lack specialized symbol manipulation, potentially limiting its capabilities.
Our neuro-symbolic pairs framework provides a solution by allowing seamless translation between paradigms. 
For instance, developers could use sparse neural approaches to acquire knowledge from large amounts of data---something not easily accomplished using symbolic methods---and then translate this neural model into a symbolic system that offers advanced symbol manipulation capabilities and the potential for incorporating additional hand-authored knowledge. 

Examples of neuro-symbolic pairs already exist in the literature.
For example, logistic regression can be viewed as a neural network without a hidden layer, while na\"{i}ve Bayes, which learns a prototype for each label, represents a simple symbolic system.
These approaches form a neuro-symbolic pair because \citet{ng2001discriminative} demonstrated that they have equivalent representations and that na\"{i}ve Bayes models can be translated into comparable logistic regression models.
Another example comes from \citet{pmlr-v108-silva20a}, who explore differentiable decision trees. Their work establishes a neuro-symbolic pair for univariate decision tree learning, where they use a neural network to learn a decision tree and then translate it into a symbolic system for use in reinforcement learning.
Although both of these prior works provide examples of neuro-symbolic pairs, they do not describe their work in these terms.
We argue, however, that the concept extends far beyond these early examples and has significant broader potential. 


\section{A Neuro-Symbolic Pair for Taxonomic Nets}

Taxonomic networks are a type of discrimination network in which nodes represent taxonomic categories arranged hierarchically based on shared attributes. These networks facilitate efficient categorization and generalization by structuring knowledge in a tree-like format, where broader concepts progressively refine into more specific subcategories---mirroring human concept organization \citep{corter1992explaining}.

\subsection{Symbolic Instantiation of Taxonomic Nets}

A classic symbolic approach to learning taxonomic networks is Cobweb~\citep{fisher1987knowledge}, an incremental method that dynamically partitions data.
Unlike clustering algorithms with a fixed number of categories, Cobweb continuously refines its hierarchy, forming prototypes adaptively. %
Recent extensions of Cobweb have demonstrated its effectiveness in continual learning and low-data scenarios.
For example, we developed Cobweb/4V \citep{barari2024incremental} to support incremental formation of visual concepts, and showed that it can achieve performance comparable to neural networks while being more robust to catastrophic forgetting during continual learning.
Similarly, we developed Cobweb/4L \citep{lian2024incremental} to support efficient language learning.
Our approach efficiently acquires word representations, outperforming several neural methods, even with significantly less training data.
These findings underscore the potential of symbolic approaches for taxonomic networks.

\subsubsection{Representation}

Our prior approach \citep{maclellan2022efficient,barari2024incremental,lian2023cobweb,lian2024incremental} represents concepts using \textit{probabilistic prototypes}, where each node in the hierarchy encodes the statistical properties of all instances categorized under it.
We assume that each prototype is normally distributed with independent features.
To track these distributions, each concept node $c$ maintains mean ($\mu_{c}$) and variance ($\sigma^2_{c}$) vectors, which are incrementally updated as new instances are assigned to the concept.

\subsubsection{Performance}


To categorize an instance $x$, the system performs a best-first search up to $n$ nodes.
Starting from the root, it expands the $n$ nodes from the taxonomy that best represent the instance and have the most predictive power.
At each search step, it selects and expands the node $c^*$ with the highest \textit{collocation score}, defined as $s(c) = p(c|x)p(x|c)$ \citep{jones1983identifying}. Letting $\mathcal{C}^*$ represent all the nodes expanded during categorization, Cobweb estimates the probability of each attribute $x_i$ as the collocation-weighted mixture of the expanded nodes' stored probability distributions:
\[
    p(x_i \mid \mathcal{C}^*) = \sum_{c\in\mathcal{C^*}}p(x_i \mid c)\frac{\exp\{s(c)\}}{\sum_{c\in\mathcal{C^*}}\exp\{s(c)\}} \label{eqn:predict-value}
\]

\subsubsection{Learning} 
To update the hierarchy, each new training instance $x$ is categorized down the tree.
At each branch, our approach considers four possible operations to update the hierarchy: (1) \textbf{adding} the instance to one of the existing children, (2) creating a new node that \textbf{merges} two of the children and inserting the instance into the merged node (the original children become children of the new node), (3) \textbf{splitting} the concept that best matches the instance and promoting its children, and (4) creating a \textbf{new} child to store the instance. 
The system chooses the operation that maximizes the 
Kullback–Leibler divergence ($D_{KL}$) between the probability distributions stored at the parent concept ($c_{\textit{parent}}$) and each child concept ($c_{\textit{child}}$) according to the following formula:
\[
\sum_{\textit{child}}  p(c_{\textit{child}}) D_{KL}\left(p(x|c_{\textit{child}}) \parallel p(x|c_{\textit{parent}})\right)
\]

\noindent where $p(x|c) \sim \mathcal{N}(\mu_{c}, \Sigma_{c})$ and $\Sigma_c = \text{diag}(\sigma_c^2)$, under the assumption that the features are independent and normally distributed.
In lay terms, this measure maximizes the information gained by knowing the concept label $c$ for an instance over the label of its parent. 

\subsection{Neural Instantiation of Taxonomic Nets}
To construct a neural-symbolic pair, we developed a novel neural architecture that has representational equivalence with our symbolic approach.
It uses a neural network that is organized in a tree structure, where each neuron corresponds to a taxonomic concept.

\subsubsection{Representation}
Neural taxonomic nets encode both a gating function \(g_\theta(x)\in[0,1]\) and a linear layer that approximates the class distribution \(p_\phi(y|x)\) at each node. Following prior work on neural soft decision trees \citep{HMOEs,Frosst2017-ed,Irsoy2014-ks,Wan2020-li}, \(g_\theta(x)=\sigma(x\mathbf{W}+\mathbf{b})\) is a linear layer with parameters \(\mathbf{W}\) and \(\mathbf{b}\) followed by a sigmoid activation that encodes the left branch probability. Consequently, the right branch probability can be inferred as \(1-g(x)\).
While we focus on binary trees in our work, this architecture can be extended to trees with branching factor \(>2\) by replacing the sigmoid with a softmax function. 

\begin{figure}
    \centering
    \includegraphics[width=0.8\linewidth]{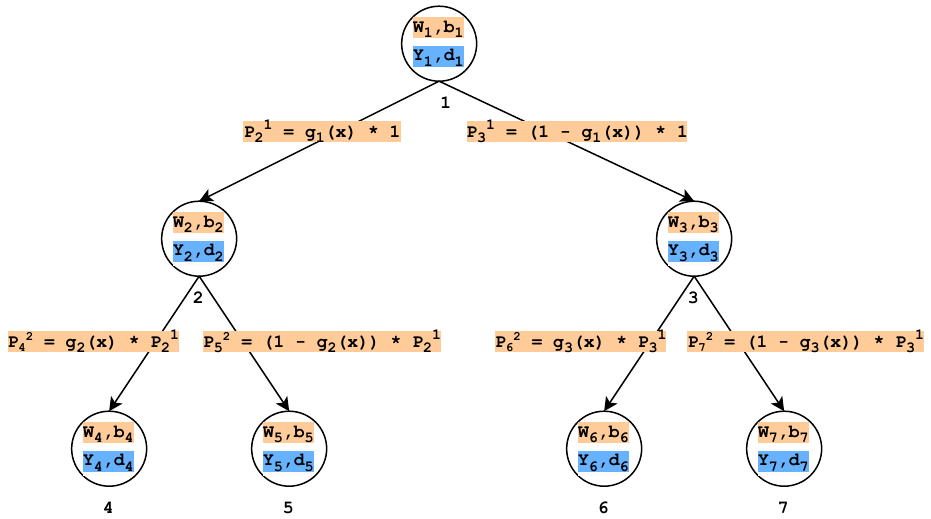}
    \caption{Neural taxonomic net with three levels. Path probability \(P^l_c\) that \(x\) arrives at node \(c\) at level \(l\) and its weights are highlighted in yellow. Weights for classification are highlighted in blue. \vspace{-10px}
    }
    \label{fig:cobwebNN}
\end{figure}




To control the smoothness of gating decisions, we introduce a temperature \(\tau\in(0,\infty)\) such that \(\tau=1\) gives the original sigmoid and \(\tau<1\) approximates the step function.
To support categorical decisions (i.e., \(g_\theta(x)\in\{0,1\}\)) while allowing proper gradient flow, we use the straight-through trick \citep{jang2017categoricalreparameterizationgumbelsoftmax}.
We also introduce stochasticity in the gating function to avoid greedy paths and encourage the exploration of other branches.
Following \citet{jang2017categoricalreparameterizationgumbelsoftmax}, we add a small Gumbel noise \(G\) scaled by \(\alpha\) that controls the strength of the noise to the output from the linear layer. As a result, the probability of taking the left branch given \(x\) at node \(c\) is:

\[
p_c(x) = \sigma\left(\frac{({x\mathbf{W_c}+\mathbf{b_c}}) +\alpha G}{\tau}\right).
\]

We can further define the path probability \(P^l_c(x)\) as the probability that \(x\) reaches a specific node \(c\) at level \(l\) in the tree.
\(
P^l_c(x)=p_c(x)\cdot P^{l-1}_\textit{parent}(x),
\)
where \(P^0_\textit{root}(x)=1\). Figure \ref{fig:cobwebNN} shows an example of a three-layer neural taxonomic net with path probabilities.

Finally, the class distribution at each node \(c\) is parametrized by a linear layer that maps from the feature space (\(H\)) to the class space (\(K\)):
\(
p_c(y|x) = x\mathbf{Y_c}+\mathbf{d_c},
\)
where \(\mathbf{Y_c}\in\mathcal{R}^{H\times K}\) and \(\mathbf{d_c}\in\mathcal{R}^{K}\).

\subsubsection{Performance}
Neural taxonomic nets leverage the entire tree to make predictions. At each level \(l\), the tree combines \(p_c(y|x)\) for each node \(c\) at that layer weighted by its path probability \(P^l_c(x)\). In the categorical setting, prediction will be based on a single path, where only the nodes along a path given \(x\) are used. For all nodes at level \(l\) and for each level of the tree:
\[
p(y|x) = \sum_l\sum_{c\in l}P^l_c(x)\cdot p_c(y|x)
\]

\subsubsection{Learning}

We train neural taxonomic nets end-to-end using gradient descent and back-propagation to update the gating functions and classifiers. The learning objective is the sum of negative log-likelihood of \(p_c(y|x)\) at each node, weighted by path probabilities. Formally, the loss function \(\mathcal{L}\) is given by:
\[
\mathcal{L}_{CE}(x,y) = \sum_{l}\sum_{c\in l} P^l_c(x)\cdot\left[-\log\frac{\exp\big(\ell_{c,y}(x)\big)}{\sum_{k=1}^{K}\exp\big(\ell_{c,k}(x)\big)}\right],
\]
where \(\ell_{c,k}(x)\) is the probability for class \(k\) at node \(c\):
\(
\ell_{c,k}(x) = \big[x\mathbf{Y}_c+\mathbf{d}_c\big]_k.
\)

To avoid trivial decisions that send all examples down a single path, we add a regularization term that encourages splitting at each node, similar to an approach by \citet{Frosst2017-ed}. 
The regularization is the KL divergence between the predicted activation distribution $A_l(c)=\texttt{softmax}({\sum_x P^l_c(x)})$ and the uniform activation distribution $U_l(c)= 2^{-l}$ at the layer \(l\): $C=\sum_l{D_{KL}}\left(A_l \parallel U_l\right)$.
\noindent The final loss function is $\mathcal{L}=\mathcal{L}_{CE}-\lambda C$, where \(\lambda\) weights the strength of the regularizer.



\subsection{Translating Between Approaches}\label{sec:translate}

These two approaches form a neuro-symbolic pair because it is possible to translate a model from one approach into an equivalent model in the other. 
For simplicity, let’s assume the taxonomic nets are binary, corresponding to our earlier descriptions.
Since each branch in the symbolic framework is essentially a na\"{i}ve Bayes classifier, there is a direct mapping to the corresponding gating function $g_\theta(x)=\sigma(x\mathbf{W}+\mathbf{b})$.
In particular, $W = \frac{(\mu_{\textit{left}} - \mu_{\textit{right}})}{\sigma_{\textit{parent}}^2}$ and $b = \ln \frac{p(\textit{left})}{p(\textit{right})} + \frac{(\mu_{\textit{right}}^2 - \mu_{\textit{left}}^2)}{2 \sigma_{\textit{parent}}^2}$.
Similarly, the classification distribution, $p_\phi(y|x)$, at each node $c$ is set to $p(y|c)$.

\begin{figure}[t]
    \centering
    \includegraphics[width=1\linewidth]{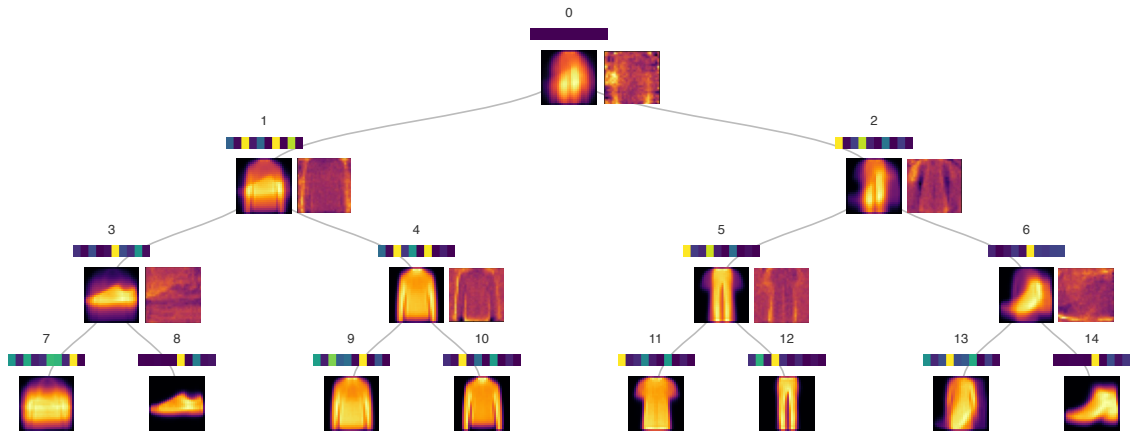}
    \caption{A three-level neural taxonomic net trained on FashionMNIST. The right image at each node shows the learned gating weights, and the left image displays the average of all test samples that pass through it (prototypes). The color bar on the top of each node shows the learned class label distribution (the root node is uniform because there are an equal number of examples in each class). 
    }
    \label{fig:enter-label}
\end{figure}

The reverse direction is not as straightforward because the neural decision tree does not store the distributional information ($\mu_c$ and $\sigma_c$) for each node $c$.
As a result, there are an infinite number of symbolic models that correspond to a particular neural model---each corresponds to a symbolic model with centroids that are different distances from the separating hyperplane that divides them.\footnote{The decision boundary is the hyperplane that is orthogonal to the line between the two children's centroids and equidistance from each centroid.} 
To identify the best translation, we start by classifying all the data using the neural approach.
We then choose the parameters at each branch such that the prototype centroids (the $\mu$s) best align with the average of all the data points assigned to each node while still being consistent with corresponding neural decision boundary.
We set the variances (the $\sigma^2$s) to correspond to the sample variance for all the points classified under each neural node.

\section{Experiments}

\subsection{Datasets}
We evaluate taxonomic networks instantiated within both the symbolic and neural frameworks using three datasets of increasing complexity and dimensionality: MNIST, FashionMNIST, and CIFAR-10. MNIST contains 70,000 \(28 \times 28\)-pixel gray-scale handwritten digits, providing a low-dimensional dataset to assess clustering proficiency with minimal feature overlap. FashionMNIST follows the same data format as MNIST, but contains everyday clothing objects that have more complex features and intra-class variations. CIFAR-10 contains 60,000 \(32 \times 32\)-pixel color images of everyday object classes. All three datasets have 10 labeled classes from which we reserve 10,000 images for testing. 



\subsection{Methods}

For learning taxonomic nets within the symbolic framework, we process every instance individually (i.e. batch size \(=1\)), with one-hot class labels imprinted on the first 10 pixels of each image \citep{hinton2022forwardforwardalgorithmpreliminaryinvestigations}. During prediction, we collect the first 10 pixel values as the predicted class label distribution.
When using the neural framework, we use batch learning, so we utilize batches of 128 and initialize the tree with 8 layers.
During training, we use the Adam \citep{kingma2017adammethodstochasticoptimization} optimizer with a learning rate of $2 \times 10^{-3}$. We also use the following hyper-parameters: $\tau = 0.3,\:\alpha=0.3,\:\lambda=110$, which were identified with hyper-parameter search. We run each experiment 8 times, each with a different random seed.
Within each run, we set the number of epochs to 10.


\subsection{Results}
\subsubsection{Comparison of Accuracy}

The symbolic approach finds the best taxonomic nets on MNIST with an average accuracy of 96.42\%.
However, our neural approach finds better taxonomic nets in FashionMNIST and CIFAR-10 with accuracies of 86.72\% and 37.95\% respectively. We see the neural approach better handles more complex datasets like FashionMNIST and CIFAR-10 compared to it's symbolic counterpart.



\begin{table}[]
    \centering
    \small
    \begin{tabular}{|c|c|c|c|}
    \hline
    \textbf{Approach $\backslash$ Data} & \textbf{MNIST} & \textbf{FashionMNIST} & \textbf{CIFAR-10} \\
    \hline
    Symbolic Learning       
       & \textbf{96.42\%$_{\pm0.06\%}$}
       & 84.04\%$_{\pm0.08\%}$ 
       & 33.73\%$_{\pm0.07\%}$ \\
    Neural Learning     
       & 96.29\%$_{\pm0.03\%}$ 
       & \textbf{86.72\%$_{\pm0.08\%}$}
       & \textbf{37.95\%$_{\pm0.76\%}$} \\

    \hline
    \end{tabular}
    \caption{Model accuracies with standard errors computed from 8 random seeds across three datasets.
    }
    \label{tab:combined-results}
\end{table}

\subsubsection{Comparison of Learning Curves}
\begin{figure}[ht]
    \centering
    \includegraphics[width=0.8\linewidth]{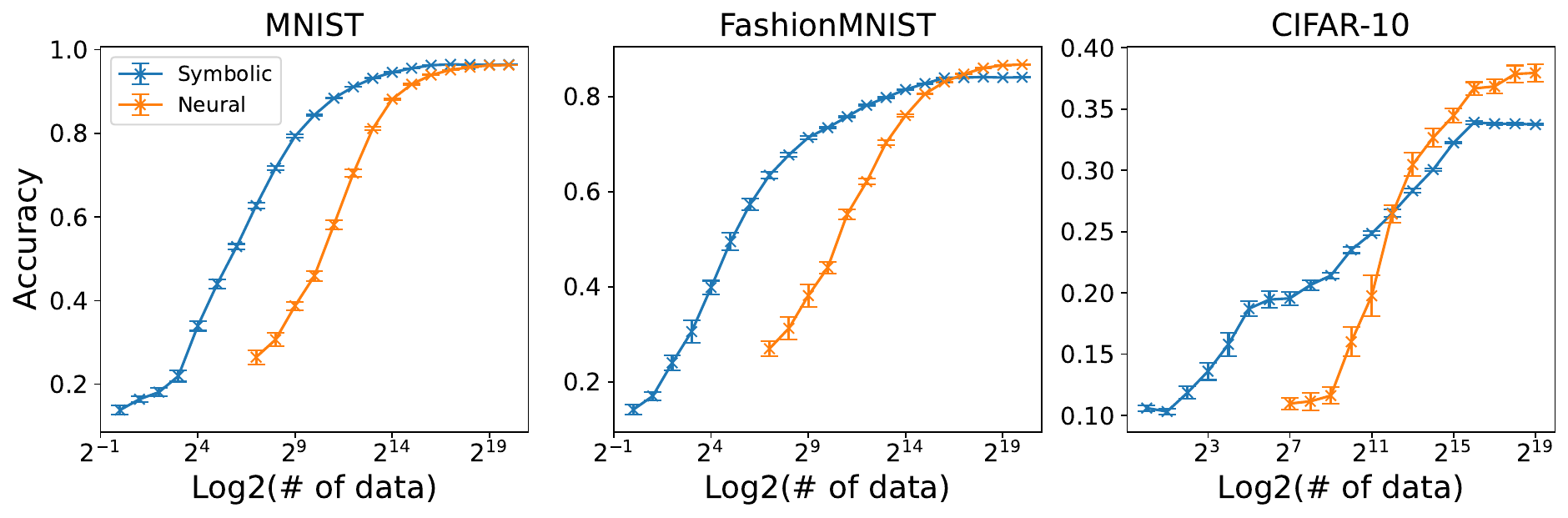}
    \caption{Learning curves for the symbolic and neural approaches. Accuracies are averaged over 8 random seeds with standard errors. 
    }
    \label{fig:learning-curves}
\end{figure}

To investigate the learning dynamics between the symbolic and neural approaches, we plot their learning curves on each dataset in Figure \ref{fig:learning-curves}.
For each approach, we fix the ordering of the training data using a random seed and record its test accuracy at every power of two data points.
The neural approach starts at $2^7$ data points because its batch size is 128.
Our results comply with the previous findings that a generative approach (comparable to our symbolic method) will have better performance with fewer data than a discriminative approach (comparable to our neural method) while reaching worse asymptotic accuracy when more examples are provided \citep{disc_vs_gen}.

\subsubsection{Comparison of Compute, Run Time and Model Memory Footprint}

We evaluated the compute efficiency of symbolic and neural approaches across compute resources, wall time, and memory usage
Our symbolic approach learns incrementally, one datum at a time, making it well-suited for CPUs and benefiting from unified memory architectures like Apple’s M4 Pro for tree-based operations, such as merging and splitting.
It ran on a single CPU core, while the neural approach used an NVIDIA A40 GPU.
Across the three datasets, the symbolic approach took 86.05s on MNIST, 89.46s on FashionMNIST, and 233.80s on CIFAR-10. In contrast, the neural approach took 133.14s, 130.58s, and 168.88s, respectively. These times are averaged over five runs.
The neural approach scales better with higher-resolution images in CIFAR-10 due to GPU acceleration, whereas the symbolic approach, running exclusively on a single CPU, faced scalability limitations.
For memory usage, the symbolic approach peaked at around 700MB, while the neural approach (with an 8-layer taxonomic net and a batch size of 128) peaked at around 500MB.

\section{Discussion}

Our results highlight several trade-offs between our two approaches, showing that one is not strictly better than the other.
Across all three datasets, the symbolic approach is more data-efficient, achieving higher accuracy with less data.
While the neural approach is less data-efficient, it finds higher-performing taxonomic nets.
This mirrors prior research on generative-discriminative pairs.
\citet{ng2001discriminative} found that generative approaches achieve their asymptotic performance with less data, but discriminative approaches tend to perform better with more data.
Our symbolic approach is generative because it learns the distributional form of the prototypes, and the neural variant is discriminative because it learns the decision boundaries at each branch.
Our results suggest that this prior work generalizes to more complex models like taxonomic nets.

The symbolic approach is more computationally efficient because it selectively manipulates its internal symbols.
For example, during learning, it sorts each image down the tree and only updates the nodes along a single categorization path, leaving other nodes untouched and avoiding unnecessary computation.
Similarly, during inference, it utilizes best-first search to expand only the most relevant portions of the tree.
In contrast, the neural approach performs full computation at every node during both learning and inference.
While this is less efficient, it compensates through hardware scalability.
Its neural framework makes it possible to leverage  GPUs for both training and inference, letting it train on larger images (CIFAR) using less wall time than the symbolic approach.
We have explored parallelizing the symbolic variant, but its commitment to discrete choices (e.g., which branch to choose) makes it difficult to translate onto tensor processing hardware.

Given these tradeoffs, it is fortunate that these approaches form a neuro-symbolic pair, as we can translate between paradigms to suit our needs---a pair that represent the same model.
For example, the symbolic approach is more efficient during training (less data and compute needed), but it is harder to scale up  because it cannot leverage GPUs.
Therefore, we can learn a taxonomic net more cost effectively using the symbolic approach, then translate it (see Section~\ref{sec:translate}) into a neural model for scalable inference during deployment.
Alternatively, we might imagine using the neural approach to learn high-performing taxonomic nets from large amounts of data and then translate them into symbolic models that can learn online without catastrophic forgetting \citep{barari2024incremental}.

Central to our approach is the taxonomic network representation, which enforces semantics that correspond to human symbols; nodes represent hierarchically organized taxonomic concepts.
By linking our two approaches via taxonomic networks, we aim to realize mechanistic interpretability.
Nodes in a taxonomic tree are monosemantic by design.
During inference, each example is categorized down the tree, primarily activating only a single node (or a few nodes) in each layer.
As Figure~\ref{fig:enter-label} shows, nodes represent taxonomic prototypes at increasing levels of specificity, with clear categories, such as shirt, pants, and shoes, developing at intermediate levels.
We argue this structure will result in more interpretable concepts, even when learned using data-driven, neural methods.
\vspace{-5px}

\section{Conclusions and Future Work}
Rather than focusing on the integration of distinct neural and symbolic components, our work seeks a more fundamental unification of these paradigms.
To achieve this goal, we introduce the concept of neuro-symbolic pairs.
These are linked neural and symbolic approaches that share a common knowledge representation, making it possible to translate models from one paradigm into the other.
We introduce taxonomic networks, tree-based networks where each node corresponds to a taxonomic category, and present a novel neuro-symbolic pair that utilizes these networks.
We evaluate the performance characteristics of the pair and find that each approach works best under different circumstances.
Fortunately, our pair-based framework enables translation across approaches, so we can realize the best characteristics of both.
We believe that our neuro-symbolic pairs concept is broadly applicable.
For example, other approaches, such as statistical relational learning \citep{de2010statistical,de2020statistical}, could be potentially framed in terms of neuro-symbolic pairs.
Looking into the future, we are interested in extending our pair for taxonomic networks to better support compositionality and representation learning; e.g., using processing techniques like convolution \citep{maclellan2022convolutional}.
We hope that our work inspires the development of additional pairs, such as matched neuro-symbolic variants for hierarchical task planning, and lays the foundation for new types of neuro-symbolic computation.

\acks{
This work was supported in part by the ARL STRONG Program (\#W911NF2320203). The views and conclusions contained herein are those of the authors and should not be interpreted as representing the official policies or endorsements of the sponsoring agency.
We thank Pat Langley, Doug Fisher, Jesse Roberts, Kyle Moore, Xin Lian for providing feedback on this draft and on our ideas that grew into this work.}

\bibliography{ref}



\end{document}